\documentclass[a4paper]{article}

\usepackage[utf8]{inputenc}
\usepackage{graphicx}
\usepackage{amsmath}
\usepackage{url}
\usepackage{verbatim}

\begin{document}

\title{Mislabel Detection of Finnish Publication Ranks}

\author{Anton Akusok$^1$, Mirka Saarela$^2$, Tommi K\"{a}rkk\"{a}inen$^1$,\\ Kaj-Mikael Bj\"{o}rk$^3$, Amaury Lendasse$^{4,5}$}
\date{%
    $^1$Arcada University of Applied Sciences, Helsinki, Finland\\
    $^2$University of Jyv\"{a}skyl\"{a}, Jyv\"{a}skyl\"{a}, Finland\\
    $^3$Risklab at Arcada UAS, Helsinki, Finland\\
    $^4$Department of Mechanical and Industrial Engineering, \\The University of Iowa, Iowa City, USA
    $^5$The Iowa Informatics Initiative, The University of Iowa, Iowa City, USA
}

\maketitle

\begin{abstract}

The paper proposes to analyze a data set of Finnish ranks of academic publication channels with Extreme Learning Machine (ELM). The purpose is to introduce and test recently proposed ELM-based mislabel detection approach with a rich set of features characterizing a publication channel. We will compare the architecture, accuracy, and, especially, the set of detected mislabels of the ELM-based approach to the corresponding reference results in \cite{saarela2016expert}.

\end{abstract}

\section{Introduction}

Finland, in the spirit of Norway and Denmark, introduced ranking system for academic publication channels (referring to scientific journals, conference series, book publishers etc.) called as \emph{Jufo} (i.e. "Julkaisufoorumi" in Finnish, "Publication Forum" in English) in 2010, together with the renewed university legislation. The ranking of a publication channel, ranging from 0 (non-peer-reviewed) to 3 (most distinguished academic publication forums), is decided by a specially nominated panel of a particular scientific discipline. These panels decide the rankings based on their academic expertise in regular meetings. Because the rankings are directly linked to the allocated funding of the universities, there has been and is a lot of discussion about the fairness and objectivity of the ranks.

A versatile analysis of the 2015 Jufo-rankings was done in \cite{saarela2016expert}. There, by using association rule mining, decision trees, and confusion matrices with respect to Norwegian and Danish ranks, it was shown that most of the expert-based rankings could be predicted and explained with machine learning methods. Moreover, it was found out that those publication channels, for which the Finnish expert-based rank is higher than the estimated one, are characterized by higher publication activity or recent upgrade of the rank. Hence, the outcomes of the system, the publication ranks, need to be assessed and evaluated regularly and rigorously.

Extreme Learning Machine (ELM), as proposed by Huang et al. \cite{huang2006extreme,huang2012extreme}, provides one of the key randomized neural network frameworks \cite{gallicchio_randomized_2017}. Probabilistic convergence analysis of the technique was provided in \cite{liu2015extreme,lin2015extreme}, where the necessity of repeated sampling of the feedforward kernel and the advantage of weight decay (ridge regression) were concluded. Here, to identify possibly mislabeled publication channel ranks, we apply the MD-ELM algorithm described and successfully tested in \cite{akusok2015md}.

The rest of the paper is organized as following. The next section~\ref{sec:data} introduces the original dataset of \emph{Jufo} rankings. The methodology, section~\ref{sec:methodology}, describes the feature extraction process and summarizes the MD-ELM method. Section~\ref{sec:experiments} explains the experimental setup, general prediction performance, and provides the comparison with the previous results in~\cite{saarela2016expert}. The last section~\ref{sec:conclusions} summarizes the findings and describes the future research directions.

\section{Data}
\label{sec:data}

The data for this study comes from two publicly available databases containing the 
Finnish publication source information and the actual national publication activity information. 
\begin{enumerate}
\item{\emph{JuFoDB}:} database of the Finnish publication forum, 
\emph{JuFo}\footnote{Available at \url{http://www.tsv.fi/julkaisufoorumi/haku.php}.}, 
which contains all nationally evaluated publication channels. Data was retrieved from this database in February 2015, so it describes the ranking situation after complete reevaluation round by the end of 2014.
\item{\emph{JuuliDB}:} The publicly accessible database of \emph{Juuli}\footnote{Available at \url{http://www.juuli.fi/?&lng=en}.} 
that contains all publications of Finnish researchers. Each
publication channel in \emph{JuFoDB} has a unique \emph{Juuli} ID, through which all Finnish publications in that particular
channel can be found. Data was retrieved from this database in September 2015, because only then all published work by the end of 2014 had been checked and included in the repository. 
\end{enumerate}

29,443 different publication channels with 33 attributes were retrieved from \emph{JuFoDB} and 107,289 publications from 
\emph{JuuliDB}. 
The Finnish expert-based \emph{rank} of each publication channel as well as  
the Norwegian and Danish expert-based rankings can be obtained directly through the \emph{JuFoDB} and also   
the three bibliometric indicators from Scopus, that is the \emph{SJR}, the \emph{SNIP} and the \emph{IPP}, are featured.
Moreover, through the link to \emph{JuuliDB}, one can directly access the information of all researchers in Finland who have published in the particular channel.

The \emph{panel} variable determines the list of experts\footnote{See \url{http://www.julkaisufoorumi.fi/en/publication-forum/panels}.} who have evaluated the publication channel and decided the Finnish expert-based rank. It basically 
indicates the research discipline of the publication channel. 
\emph{Field}, \emph{MinEdu field}, \emph{Web of Science fields}, \emph{Scopus fields} are further variables indicating the discipline 
of the publication channel. However, multiple linkings are possible for these variables and for some publication channels these linkings are not available at all. But each publication channel is attached to only one panel and the panel information is available for all publication channels except for 6,562 book publishers that have mostly been evaluated as rank 0 \cite{saarela2016expert}.

In addition to some more general data, such as the \emph{title}, \emph{subtitle}, \emph{website}, 
\emph{country of publication}, \emph{language}, unique identifier (\emph{ID}), \emph{ISSN}, \emph{Sherpa/Romeo code}, \emph{starting year}, and \emph{publisher}, 
the \emph{JuFoDB} also provides information such as \emph{abbreviation}, \emph{title details}, \emph{ISBN}, \emph{DOAJ}, \emph{end year}, \emph{continued under the name}
and \emph{continued JuFo-rank}. 
The \emph{evaluation history} provides information about the previous ranks in the system. 

Similarly as in \cite{saarela2016expert}, the continuous variables are directly utilized as features and the 
categorical variables are transformed to own binary features for each category. 
All of the 29,443 publication channels have missing values 
for at least some of the 33 total variables. Hence, for utilizing all of available data in the analysis, one faces a significant sparsity problem \cite{MITstudentPaper}.
Since the missing information was discovered as an important predictor of the Finnish expert-based rank in \cite{saarela2016expert}, 
we utilize here all the described variables as features plus for each variable the binary information whether it has an available value. 
Thus, for our final model we had 942 features (452 original + 400 added non-linear feature combinations). 


\section{Methodology}
\label{sec:methodology}

\subsection{Feature extraction}

The original variables as described in the previous section 
were transformed into numerical features, either real-valued or binary ones. Each original feature has its own specific transformation into numerical format. The absence of a value, similarly to \cite{saarela2016expert}, is encoded with a separate binary variable for most features, as it provides valuable information (i.e., absence of a website of a poor quality conference). 

The original features that are used for the analysis task, and their corresponding transformations are described below in Table~\ref{tab:features}. The results are notably missing Jufo-rankings for the previous years; those are omitted on purpose to make the rank prediction task unbiased by the previous decisions.

\begin{table}[]
    \centering
    \caption{The list of original features and their numerical representations.}
    \label{tab:features}
    \begin{tabular}{p{0.03\linewidth}|p{0.13\linewidth}|p{0.30\linewidth}|p{0.50\linewidth}}
\# & Feature & Meaning & Numerical representation \\
\hline
1 & Level & Current Jufo ranking & An output variable with integer values in range $\{0,1,2,3\}$ \\
2 & Title,\newline Subtitle & Title and subtitle (if available) of the publication & Encoded in a Bag-of-Words representation, dimensionality reduced from 3700 to 30 by a Sparse Random Projection \\
3 & Website & Website of the publication & Country code of the host represented in one-hot encoding with 117 binary variables (including \emph{unknown}) \\
4 & Type & Publication type (journal, conference, book series) & Represented in one-hot encoding with 3 binary variables; this feature has no missing values \\
5 & ISSN & ISSN numbers of printed and online versions & A binary variable representing whether the publication has an ISSN, two variable total \\
6 & StartYear & Start year of the publication & A logarithm of age of the publication, plus a binary variable representing missing value \\
7 & Publication Country & Country of publication & One-hot encoding of the publication origin with 114 binary variables, including the \emph{unknown} origin \\
8 & Publisher & Publisher of the series & One-hot encoding of 100 most popular publishers, plus \emph{other} publisher \\
9 & Language & Language of the publication & One-hot encoding of the publication language with 49 binary variables, including \emph{undetermined} \\
10 & ERIH-class & ERIH ranking of publications & One-hot encoding of the four available ranks, plus a \emph{missing} rank \\
11 & SJR\newline SNIP\newline IPP & Impact factors in three different systems & Three real-valued variables for the impact factors, plus three binary variables indicating the absence of an impact factor \\
12 & DOAJ\newline Sherpa/Romeo & Open access types & Eight binary variables: two for DOAJ levels, and six for the Sherpa/Romeo levels \\
13 & Field & The field of study in Finnish classification & Ten binary variables for the ten fields, a publication may belong to multiple fields \\
14 & MinEdu Field & The field of study according to the Ministry of Education classification & 70 binary variables for the Ministry of Education fields, a publication may belong to several of them \\
15 & Panel & The scientific panel that assigned a corresponding score & One-hot encoding of the panel number with 25 binary variables, including a \emph{not available} panel \\
16 & ISBN & ISBN numbers used by the publication & One variable representing the number of different ISBNs; can be zero
    \end{tabular}
\end{table}

\subsection{Mislabel detection using MD-ELM}\label{MD-ELM}

The mislabel detection is based on the MD-ELM algorithm from \cite{akusok2015md}. The key idea is to include in a data set artificial mislabels, which then can be used as baseline in a statistical detection of unknown mislabels using Welch's t-test and directly computable Leave-One-Out (LOO) cross-validation error (PRESS statistics). In this way, the MD-ELM algorithm detects samples whose original labels are likely incorrect.

More precisely, the MD-ELM analyses the changes in the LOO error of the model in response to randomly changing labels of a few training samples. If the new labels reduce the global LOO error, the mislabel score of those samples is increased. A small part of the samples, whose labels are randomly changed on purpose, create the control group called \emph{artificial mislabels}. Scores of the artificially mislabeled samples help to determine whether the MD-ELM method succeeds, and define the stopping criterion.

The mislabel detection method uses Extreme Learning Machine as the powerful nonlinear prediction model with a fast LOO error. A practical implementation employs several ELM models with different sets of artificial mislabels, eliminating their possible impact on the results. The predicted originally mislabeled samples are samples with the mislabel score higher that the given quantile of a normal distribution fitted to all the scores.

\section{Experimental results}
\label{sec:experiments}

\subsection{Prediction performance}

A successful MD-ELM method application requires a precise prediction model to work with. The prediction task uses features 2-16 from Table~\ref{tab:features} as inputs and the feature 1 as the target output. 

The dataset exhibit a strong class imbalance (see Table~\ref{tab:imbalance}). The imbalance causes rank 3 to be completely neglected in the predictions, unless class balancing measures are taken.

\begin{table}[]
    \centering
    \caption{Number of data samples of each Jufo-rank.}
    \label{tab:imbalance}
    \begin{tabular}{l|c|c|c|c}
         Rank & 0 & 1 & 2 & 3 \\ \hline
         \# & 5,743 & 20,503 & 2,329 & 668
    \end{tabular}
\end{table}

\begin{figure}[ht]
    \centering
    \includegraphics[width=0.5\textwidth]{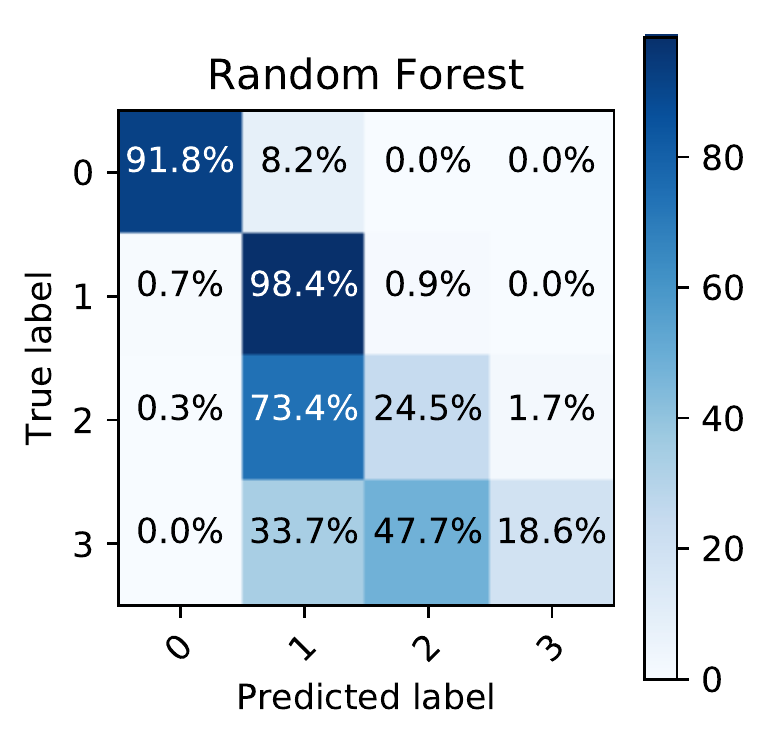}
    \caption{Confusion matrix of Random Forest classifier on out-of-batch data.}
    \label{fig:rf}
\end{figure}

The benchmark performance level is obtained with the Random Forest classifier. It achieves 89.3\% test accuracy, but the predictions are biased due to the strong class imbalance as shown on Figure~\ref{fig:rf}. The smallest class 3 has only 18,6\% correct predictions, while the largest class 1 is predicted correctly 98,4\% of times.

Unfortunately, Random Forest model cannot be used in the Mislabeled Detection framework. So an Extreme Learning Machine was train instead. The input features consisted of the 542 numerical features derived from the data, 200 standard non-linear ELM neurons and another 200 Radial Basis Function neurons. 

The output layer training proved difficult due to both class imbalance, and a high number of irrelevant linear features. The only successful model was an ElasticNet linear classifier that combined L1 and L2 regularization, trained with the Stochastic Gradient Descend. The regularization strength parameter is found by a 5-fold stratified cross-validation, that keeps the proportion of samples from different classes equal between the folds. Additionally, the method performed class balancing by computing the corresponding sample weights.

\begin{figure}[ht]
    \centering
    \includegraphics[width=0.5\textwidth]{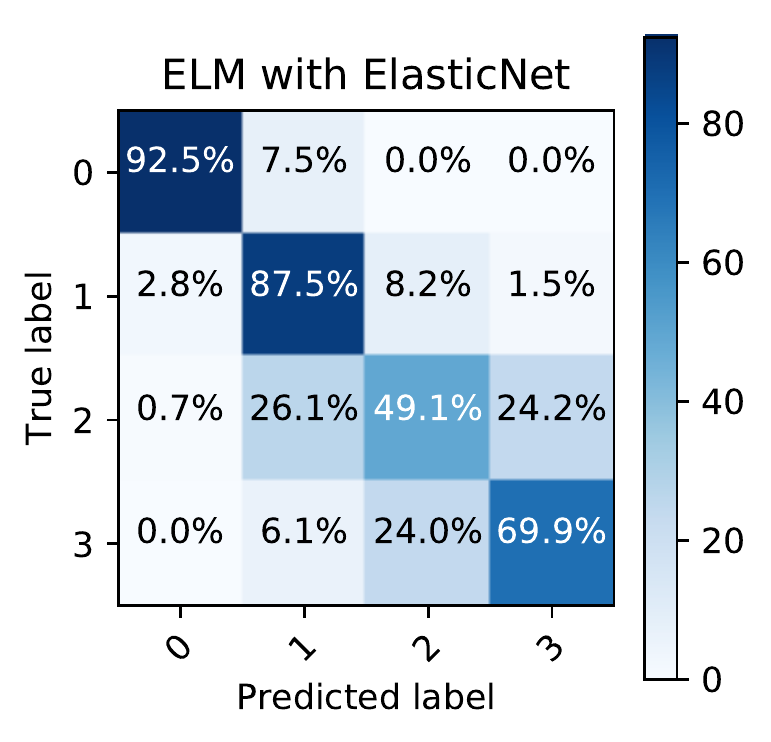}
    \caption{Confusion matrix of an ELM model with an ElasticNet classifier output layer.}
    \label{fig:elm2}
\end{figure}

The resulted ELM achieved 85\% total accuracy, distributed much more equally among the classes as shown in Figure ~\ref{fig:elm2}. The resulting model selected only 289 input features out of the total of 942, reducing the data size for the MD-ELM method.

\subsection{MD-ELM performance}

The MD-ELM method uses 289 best features selected in the prediction experiment. The method does not implement class balancing, so the scope of the experiment is limited to detecting incorrectly labeled samples of rank 3 using a dataset of 900 random samples from ranks 0,1,2 plus all the 668 samples of rank 3. Such reduced dataset has a smaller class imbalance, that does not negatively affect the results.

The final predictions are averaged over 10 different MD-ELM models. Each model uses its own dataset with different random samples of ranks 0,1,2, a random subset of 100 input features out of the available 289, and a different random subset of 3\% artificially mislabeled samples. At each iteration of the method, two samples have their labels changes, one of which is always an original rank 3 sample. 

The method continues until artificially mislabeled samples get an average score of 100. This takes 400,000 iterations. By that time, non-artificially mislabeled samples achieve an average mislabel score of only 19 with standard deviation of 28. The difference between the scores shows that MD-ELM methods succeeds at separating artificially mislabeled samples from the rest; it means that it should also succeed in detecting the originally mislabeled samples.

\begin{figure}
    \centering
    \includegraphics[width=0.49\textwidth]{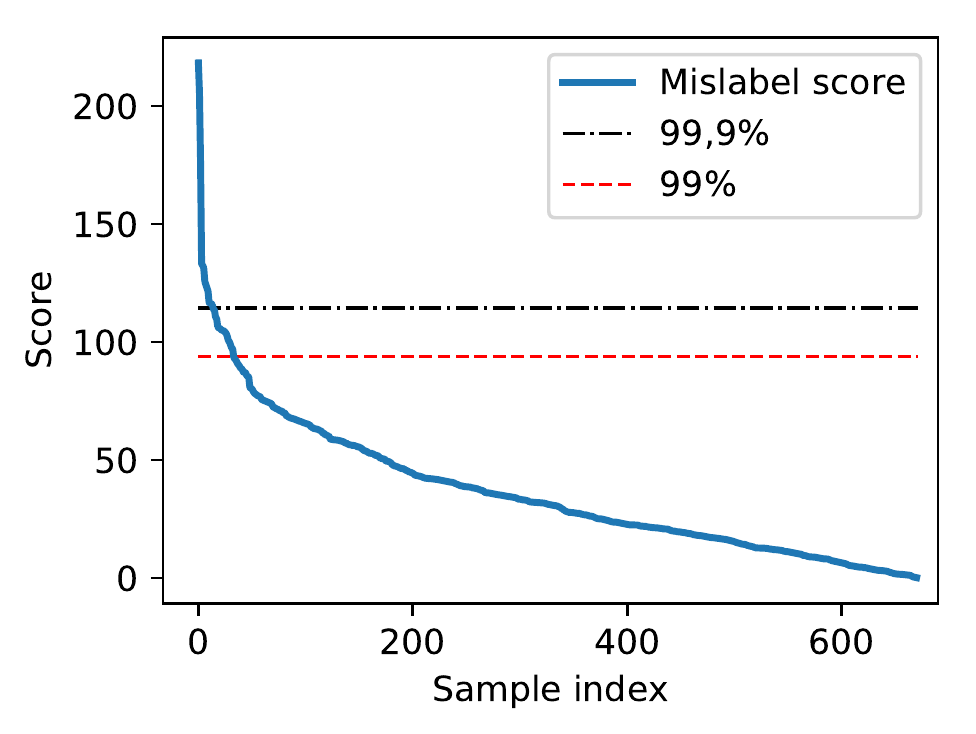}
    \includegraphics[width=0.49\textwidth]{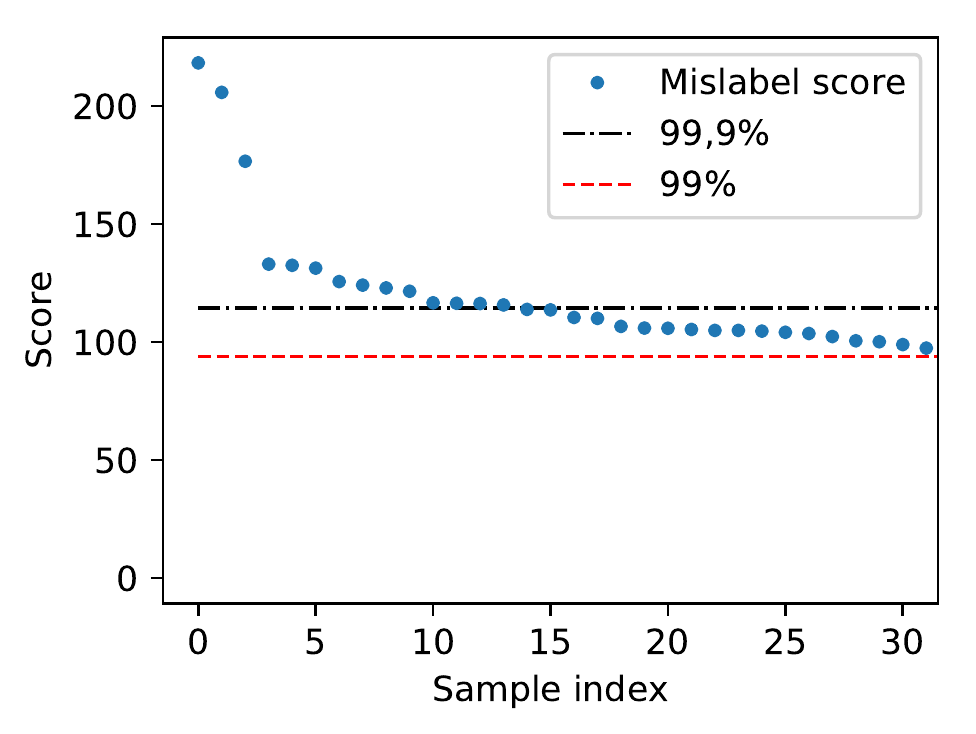}
    \caption{Mislabel scores of samples with the original rank 3 averaged over 10 MD-ELM models; zoomed version on the right. The quantile values of 99\% and 99.9\% are shown by horizontal lines. Artificial mislabels achieve an average score of 100.}
    \label{fig:scores}
\end{figure}

The mislabel scores of all the samples with the original rank 3 are shown on Figures~\ref{fig:scores}. A few outliers are clearly visible, together with other candidates to be the originally mislabeled samples. The analysis of these samples is presented below.

\subsection{Characterization of misclassified publication channels and comparison to earlier results}\label{sec:MisclassificationComparison}

As explained above, we concentrate only on misclassifications for the highest JuFo ranking, that is 
publication channels that were evaluated by the Finnish discipline experts as 3, but for which the automatic model suggested a lower rank.  
We restrict our misclassification analysis here to this set because it also resembles the largest difference to the 
Danish and Norwegian systems that include only ranks 0, 1 and 2.

With a mislabelled score over 99\% quantile of average scores, 
34 publication channels were identified for which the Finnish expert-based ranking was 3 but 
the model suggested a different rank. 
However, 30 of these misclassifications could immediately be explained by the ranks in the Danish and Norwegian model, 
which evaluated these 
publication channels as 2, that is the highest rank in their systems.

The four remaining publication channels for which both, the automated model and the Danish and Norwegian systems,  
suggested a lower rank were \emph{LIGHT: SCIENCE \& APPLICATIONS},
\emph{Etudes classiques}, 
\emph{New German critique} (for all three of these journals, the rank has recently been updated to a higher one), and the \emph{British medical journal}.
The last one has a considerable higher publication activity: The average number of Finnish publications in JuFo rank 3 channels is 
 10.78 but the \emph{British medical journal} has a total of 26 publications. All of these journals 
were also detected to be mislabled in \cite{saarela2016expert}, but the misclassification could actually be explained. 
The three Scopus indicators had incorrectly not been included in \emph{JuFoDB} for \emph{LIGHT: SCIENCE \& APPLICATIONS} and 
the \emph{British medical journal}. These indicators could be manually found from Scopus and in both 
cases the indicators were so high that rank 3 actually seemed justified.

Although the methods utilized in here were very different from the ones utilized in \cite{saarela2016expert}, the main results 
obtained and the misclassification detected in here are to a large extend the same as the ones in \cite{saarela2016expert}.
Thus, we conclude that methodological triangulation \cite{brymantriangulation,denzin1970strategies} has strengthen our analysis results.

\section{Conclusions}
\label{sec:conclusions}

An extended version of the analysis of Finnish publication channel ranks was provided in this paper. Compared to the reference models in \cite{saarela2016expert}, we used here much more versatile set of features, with fully nonlinear ELM-based rank prediction model. The mislabel detection was based on the MD-ELM algorithm proposed in \cite{akusok2015md} and briefly recapitulated in section \ref{MD-ELM}.


In summary, the experimental results obtained and reported in Section~\ref{sec:MisclassificationComparison} are very similar to the analysis results in \cite{saarela2016expert}. In our future work, we intend to repeat the mislabel detection also for the other ranks, especially rank 2 for which the most suspicious publication channel quality misclassifications were identified in \cite{saarela2016expert} and that, as explained above, actually contain the most misclassifications. The MD-ELM method will also be extended with a class balancing mechanism, allowing it to handle the whole original dataset.

\bibliographystyle{plain}
\bibliography{elt}

\end{document}